%% file: paper.tex
\documentclass[12pt]{article}
\usepackage[a4paper, margin=1in]{geometry}
\usepackage{amsmath, amssymb, mathtools}
\usepackage{graphicx}
\usepackage{booktabs}
\usepackage{cite}
\usepackage{hyperref}
\usepackage[all]{hypcap}
\usepackage{setspace}
\usepackage{ragged2e}
\usepackage{float}
\usepackage[skip=5pt]{caption}
\usepackage{array}
\title{Extended Physics Informed Neural Network for Hyperbolic Two-Phase Flow in Porous Media } 

\author{
\begin{minipage}[t]{0.45\textwidth}
\centering
Saif Ur Rehman\textsuperscript{1,*} \\[3pt] 
{\footnotesize
\linespread{0.9}\selectfont
Dawood University of Engineering and Technology,\\[-10pt]
Karachi, Pakistan
}
\end{minipage}%
\hfill
\begin{minipage}[t]{0.45\textwidth}
\centering
Wajid Yousuf\textsuperscript{2} \\[3pt] 
{\footnotesize
\linespread{0.9}\selectfont
Texas A\&M University,\\[-10pt]
College Station, Texas, USA
}
\end{minipage}
}

\date{}  

\begin{document}
\maketitle
\renewcommand{\thefootnote}{\fnsymbol{footnote}}
\footnotetext[1]{\footnotesize Corresponding author email: saifkhanengr@gmail.com}
\renewcommand{\thefootnote}{\arabic{footnote}}

\begin{abstract}
\justifying
The accurate solution of nonlinear hyperbolic partial differential equations (PDEs) remains challenging due to steep gradients, discontinuities, and multiscale structures that make conventional solvers computationally demanding. Physics-Informed Neural Networks (PINNs) embed the governing equations into the learning process, enabling mesh-free solution of PDEs, yet they often struggle to capture steep gradients, discontinuities, and complex nonlinear wave interactions. To address these limitations, we employ the Extended Physics-Informed Neural Network (XPINN) framework to solve the nonlinear Buckley-Leverett equation with a nonconvex flux, modeling immiscible two-phase flow in porous media. The computational domain is dynamically decomposed in space and time into evolving pre-shock and post-shock subdomains, allowing localized subnetworks to efficiently learn distinct flow behaviors, with coupling enforced via the Rankine-Hugoniot jump condition to ensure physically consistent flux continuity. We compare XPINN with standard PINNs and its variants, including PINN with artificial viscosity, PINN with Welge construction, and PINN with the Oleinik entropy condition, and across all cases, XPINN consistently outperforms the other methods, accurately resolving sharp fronts and capturing the correct physical behavior. Importantly, XPINN achieves this using the simpler Adam optimizer, whereas some PINN variants require more complex or higher-order strategies such as L-BFGS to reach comparable accuracy, demonstrating that XPINN is a robust and scalable approach for challenging hyperbolic PDEs without artificial diffusion or entropy corrections. The code is available at github.com/saifkhanengr/XPINN-for-Buckley-Leverett.
\end{abstract}

\clearpage
\section{Introduction}
\label{sec:Introduction}

The accurate solution of nonlinear partial differential equations (PDEs) has long been a central problem in applied mathematics, physics, and engineering. Classical discretization-based solvers—such as finite difference, finite element, and finite volume methods have been successful for many decades, yet they often face serious limitations in problems involving steep gradients, discontinuities, or multi-scale interactions that require fine meshes and significant computational resources \cite{Deb12,Tad12}. These issues are especially pronounced in applications such as porous media flow \cite{Roj98}, turbulence \cite{Pou22}, biological transport \cite{Lag20}, and nonlinear wave propagation \cite{Pol23}, where the dynamics often involve sharp transitions, that makes the simulation prohibitively expensive with traditional schemes. These challenges have motivated researchers to develop approaches that can address the limitations of traditional solvers.

Among these, Deep neural networks (DNNs) have emerged as powerful approximators capable of representing highly nonlinear mappings through compositions of simple activation functions \cite{Goo16}. Their success in computer vision \cite{Kri17, HeK16, Kar14}, natural language processing \cite{Hin12, Sut14}, and reinforcement learning \cite{Mni15} has inspired efforts to adapt them for scientific computing. In this context, the key difficulty is that training purely data-driven surrogates generally requires vast quantities of high-fidelity data, which are rarely available for complex PDE-governed systems. To overcome this limitation, the concept of physics-informed learning was proposed, in which neural networks are trained not only on observational data but also on the governing equations themselves. The first attempts to apply neural networks for solving PDEs date back to the early 1990s \cite{Lee90, Psi92, Mea94, Isa98, Isa00}. With the advent of modern GPUs and optimization algorithms \cite{Kin14}, these ideas have been revitalized and extended. Several physics-aware learning frameworks have since been introduced: \cite{Owh15} proposed learning machines that encode structured prior information about PDE solutions. \cite{Bru16} developed the sparse identification of nonlinear dynamics framework to infer governing equations from time-series data. \cite{Lin16} incorporated neural networks into Reynolds-averaged turbulence closures, while \cite{Wan18} demonstrated physics-informed learning for turbulence modeling. \cite{Tom17} introduced convolutional architectures to accelerate Navier-Stokes solvers. These examples illustrate how combining physics and machine learning has enabled progress across diverse scientific domains.

In particular, \cite{Rai19} introduced physics-informed neural networks (PINNs), which have become a widely used approach in this area. PINNs incorporate the PDE residual directly into the training loss through automatic differentiation, effectively transforming the task of solving a PDE into an optimization problem. This approach eliminates the need for mesh discretization and provides a flexible framework for both forward problems (predicting the solution) and inverse problems (estimating unknown coefficients or operators). PINNs have since been applied to a wide variety of tasks, including but not limited to stochastic and fractional PDEs \cite{Jie18, Zha20, Pan19, Kha19}, uncertainty quantification \cite{Yan19, Zhu19, Fra23}, elastodynamics \cite{Rao20}, high-speed compressible flows \cite{Mao20}, geostatistical problems \cite{Zhe20}, vortex-induced vibrations \cite{Rai18}, and cardiovascular modeling \cite{Rai20, Kis20, Fra20}. On the theoretical side, \cite{Shi20} developed a mathematical foundation for PINNs, while \cite{Mis20, Mis201} analyzed their generalization error in both forward and inverse contexts.

Despite these successes, standard PINNs encounter difficulties when applied to large-scale or highly nonlinear problems. Training can become unstable and slow, particularly when the solution involves sharp gradients, discontinuities, or multi-scale behavior. \cite{Fuk20, Fra21, Dia21, Alm22} analyze PINNs for 1D nonlinear hyperbolic problems and shows that the vanilla PINN fails to capture sharp shocks. Similarly, \cite{Dia22, Rod22, Zha24} and \cite{Dia24} uses attention-based PINNs and Physics-Informed Deep Operator Networks (PI-DeepONets), respectively, to tackle the complex dynamics of nonlinear PDEs with dominant hyperbolic behavior, where standard PINNs struggle. Moreover, \cite{Cou23} proposed adaptive artificial viscosity approaches within PINNs to effectively capture shocks in nonlinear hyperbolic PDEs such as the Burgers and Buckley-Leverett equations. Further, \cite{Fra201} applied a physics-informed machine learning approach that integrates deep adversarial neural networks with governing physical laws to simulate and history-match two-phase flow in porous media.
In addition, several domain decomposition strategies have been proposed in the scientific community. For instance, \cite{LiK20} employed local neural networks on discrete subdomains within a variational framework, while \cite{Kha21} advanced the variational PINN approach (hp-VPINNs). \cite{Yan21} applied the hp-VPINNs framework to solve nonlinear hyperbolic PDEs with shocks and applied it to dynamic subsurface fluid-flow problems. The conservative PINN (cPINN) method \cite{Jag20} extended PINNs to conservation laws by dividing the computational domain into subregions that communicate through flux continuity. Similarly, extended physics-informed neural network (XPINN) method was introduced to provide a more general and flexible decomposition strategy \cite{Jag201}. XPINNs support irregular space-time partitions, can be applied to arbitrary PDEs, and employ simple interface conditions that avoid dependence on geometric normal. These features make them particularly attractive for problems with shocks, heterogeneous media, or dynamic interfaces. By combining multiple subnetworks across subdomains, XPINNs reduce the complexity of each local model while also enabling natural parallelization, thereby lowering training cost without sacrificing accuracy.

In this paper, we apply XPINNs to the Riemann problem characterized by a nonconvex flux function with multiple inflection points. The Buckley-Leverett equation provides an ideal test case, as it is a nonlinear hyperbolic PDE modeling multiphase flow in porous media that generates steep saturation fronts and shock-like structures \cite{Wan18, Zha24}. Such features are notoriously difficult for standard PINNs to capture, often requiring artificial diffusion, entropy corrections, or complex network architectures, and higher-order optimizers such as L-BFGS \cite{Fuk20, Fra21, Dia21, Alm22, Dia22, Rod22, Zha24}. We employ XPINNs to study the Buckley-Leverett problem, demonstrating how the combination of physics-informed learning and dynamic subdomain decomposition enables efficient training, accurate resolution of discontinuous saturation fronts, and superior performance compared to standard PINNs, even when using the simpler Adam optimizer.

The remainder of this paper is organized as follows. In Section~\ref{sec:Methodology}, we introduce the XPINN framework and highlight how it differs from standard PINNs in solving deterministic PDEs. Section~\ref{sec:Mathematical_Formulation} provides an overview of the two-phase transport model along with the governing hyperbolic PDE. The results for the transport problem, including a comparison of XPINN with standard PINNs and its variants, are presented in Section~\ref{sec:Computational_Experiments}. Finally, Section~\ref{sec:Conclusion_and_Future_Work} summarizes our findings and discusses the outcomes.


\section{Methodoloy}
\label{sec:Methodology}

Physics-Informed Neural Networks (PINNs) have emerged as a powerful framework for solving forward and inverse problems involving partial differential equations (PDEs) \cite{Law22, Tos25}. The core idea is to leverage the universal approximation capabilities of deep neural networks (DNNs) to represent the solution of a PDE, while using the governing equation itself as a regularization constraint in the loss function \cite{Rai19}. In this section, we first outline the standard PINN formulation and then detail its generalization to XPINN \cite{Jag201}, which employs domain decomposition to enhance the capacity and efficiency of solving problems with complex solutions, such as the Buckley-Leverett equation with its sharp saturation fronts.

\subsection{Physics-Informed Neural Networks (PINNs)}
Consider a general PDE of the form:
\begin{equation}
u_t + \mathcal{N}[u] = 0, \quad x \in \Omega, \; t \in [0, T].
\end{equation}
where $\mathcal{N}[\cdot]$ is a nonlinear differential operator.

A PINN approximates the solution $u(x,t)$ with a deep neural network $u_{\theta} (x,t)$, parameterized by weights and biases $\theta$. A feed-forward neural network (FNN) with $L-1$ hidden layers can be defined layer-wise as follows:
\begin{align}
\text{Input layer:} \quad & z_0 = X = (x, t), \\
\text{Hidden layer(s):} \quad & z_l = \sigma(W_l z_{l-1} + b_l), \quad 1 \le l \le L-1, \\
\text{Output layer:} \quad & u_\theta(x,t) = z_L = W_L z_{L-1} + b_L. 
\end{align}
Here, $\sigma[\cdot]$ is a nonlinear activation function (e.g., $\tanh$), $W_l \in \mathbb{R}^{N_l \times N_{l-1}}$ and $b_l \in \mathbb{R}^{N_l}$ are the weight matrix and bias vector of the $l^{\text{th}}$ layer, and $\theta = \{W_l, b_l\}_{1 \le l \le L}$ denotes all trainable parameters.

The physics is informed by defining the PDE residual using automatic differentiation (AD):
\begin{equation}
r_{\theta}(x,t) = u_{\theta,t}(x,t) + \mathcal{N}[u_{\theta}(x,t)].
\end{equation}

The shared parameters $\theta$ are then learned by minimizing a composite loss function that penalizes deviations from both the initial/boundary data and the governing PDE:
\begin{equation}
L(\theta) = \lambda_u L_u(\theta) + \lambda_r L_r(\theta).
\end{equation}

where
\begin{equation}
L_u(\theta) = \frac{1}{N_u} \sum_{i=1}^{N_u} \left| u_{\theta}(x_u^i, t_u^i) - u^i \right|^2, 
\quad
L_r(\theta) = \frac{1}{N_r} \sum_{i=1}^{N_r} \left| r_{\theta}(x_r^i, t_r^i) \right|^2.
\end{equation}

The set $\{(x_u^i, t_u^i), u^i\}_{i=1}^{N_u}$ represents the initial and boundary training data, while $\{(x_r^i, t_r^i)\}_{i=1}^{N_r}$ are collocation points sampled from the domain interior. The coefficients $\lambda_u$ and $\lambda_r$ are weights to balance the loss terms.

\subsection{Extended Physics-Informed Neural Networks (XPINNs)}
While PINNs are effective, they can struggle with problems exhibiting multi-scale features or sharp gradients, as a single network must capture the solution's behavior over the entire domain. The XPINN framework addresses this by leveraging a domain decomposition approach, offering several key advantages:

\begin{itemize}
    \item \textbf{Enhanced Representation Capacity:} Different subdomains can employ Sub-Nets (smaller PINNs) with tailored architectures (depth, width, activation functions) suited to the local solution complexity.
    \item \textbf{Parallelization:} Sub-Nets can be trained in parallel, leading to significant computational savings.
    \item \textbf{Flexibility for Complex Geometries:} XPINN can handle both regular and irregular, non-overlapping domain decompositions.
\end{itemize}

The computational domain $\Omega$ is divided into $N_{\text{sd}}$ non-overlapping subdomains such that
\begin{equation}
\Omega = \bigcup_{q=1}^{N_{\text{sd}}} \Omega_q, \quad \text{and} \quad \Omega_i \cap \Omega_j = \partial \Omega_{ij}, \quad i \ne j.
\end{equation}

Here, $\bigcup$ denotes the union of all subdomains, and $\partial \Omega_{ij}$ represents the shared boundary between subdomains $\Omega_i$ and $\Omega_j$. In each subdomain $\Omega_q$, a separate Sub-Net $u_{\theta_q}(x,t)$ is assigned to approximate the local solution. The global solution is then given by:
\begin{equation}
u_{\theta}(x,t) = \sum_{q=1}^{N_{\text{sd}}} u_{\theta_q}(x,t) \cdot \mathbf{1}_{\Omega_q}(x,t),
\end{equation}
where $\mathbf{1}_{\Omega_q}$ is the indicator function for the $q^{\text{th}}$ subdomain, defined as:
\begin{equation}
\mathbf{1}_{\Omega_q}(x,t) \coloneqq
\begin{cases}
0, & (x,t) \notin \Omega_q, \\
1, & (x,t) \in \text{interior of } \Omega_q, \\
\frac{1}{s}, & (x,t) \in \text{interface of } \Omega_q,
\end{cases}
\end{equation}
where $s$ is the number of subdomains sharing that interface point. This definition ensures a partition of unity across the domain.

\subsubsection{Loss Functions and Interface Conditions for XPINNs}
\label{sec:loss_functions_and_interface_conditions_for_xpinns}

The key to stitching the subdomains together lies in the interface conditions enforced in the loss function. For a subdomain $\Omega_q$, let $\{(x_{u_q}^i, t_{u_q}^i)\}_{i=1}^{N_{u_q}}$ be the training data points, $\{(x_{F_q}^i, t_{F_q}^i)\}_{i=1}^{N_{F_q}}$ be the residual collocation points, and $\{(x_{I_q}^i, t_{I_q}^i)\}_{i=1}^{N_{I_q}}$ be the points on its interface with neighboring subdomains $\Omega_{q^+}$.

The total loss function for the $q^{\text{th}}$ Sub-Net in the forward problem is:

\begin{equation}
\begin{aligned}
J(\theta_q) &=
\underbrace{\lambda_{u_q} L_{u_q}(\theta_q)}_{\text{Data}}
+ \underbrace{\lambda_{F_q} L_{F_q}(\theta_q)}_{\text{Residual}}
+ \underbrace{\lambda_{IF_q} L_R(\theta_q)}_{\text{Interface Residual}}
+ \underbrace{\lambda_{I_q} L_{\text{avg}}(\theta_q)}_{\text{Interface Average (Optional)}} \\[-15pt]
&\quad
\underbrace{
  \hphantom{
    \lambda_{u_q} L_{u_q}(\theta_q)
    + \lambda_{F_q} L_{F_q}(\theta_q)
  }
}_{\text{Standard PINN}} \\[-15pt]
&\quad
\underbrace{
  \hphantom{
    \lambda_{u_q} L_{u_q}(\theta_q)
    + \lambda_{F_q} L_{F_q}(\theta_q)
    + \lambda_{IF_q} L_R(\theta_q)
    + \lambda_{I_q} L_{\text{avg}}(\theta_q)
  }\hspace{4em}
}_{\text{XPINN}}
\end{aligned}
\label{eq:total_loss}
\end{equation}

where the constituent loss terms are defined as:
\begin{align}
L_{u_q}(\theta_q) &= \frac{1}{N_{u_q}} \sum_{i=1}^{N_{u_q}} 
\left| u_{\theta_q}(x_{u_q}^i, t_{u_q}^i) - u^i \right|^2, \label{eq:data_loss} \\[6pt] 
L_{F_q}(\theta_q) &= \frac{1}{N_{F_q}} \sum_{i=1}^{N_{F_q}} 
\left| F_{\theta_q}(x_{F_q}^i, t_{F_q}^i) \right|^2, \label{eq:PDE_res} \\[6pt]
L_R(\theta_q) &= \sum_{q^+} \left[
\frac{1}{N_{I_q}} \sum_{i=1}^{N_{I_q}}
\left| F_{\theta_q}(x_{I_q}^i, t_{I_q}^i) - F_{\theta_{q^+}}(x_{I_q}^i, t_{I_q}^i) \right|^2
\right], \label{eq:interface_res} \\[6pt] 
L_{\text{avg}}(\theta_q) &= \sum_{q^+} \left[
\frac{1}{N_{I_q}} \sum_{i=1}^{N_{I_q}}
\left| u_{\theta_q}(x_{I_q}^i, t_{I_q}^i) - \langle u \rangle(x_{I_q}^i, t_{I_q}^i) \right|^2
\right], \label{eq:interface_avg}
\end{align}

where the interface-averaged solution is defined as
$\langle u \rangle = \frac{u_{\theta_q} + u_{\theta_{q^+}}}{2},$
representing the average solution along the shared interface between two neighboring Sub-Nets. This averaging condition is applied only when the interface is continuous, since it inherently enforces continuity. Therefore, this term is optional and its applicability depends on whether the discontinuity lies at the interface; if so, enforcing the average would degrade the solution.

\begin{figure}[h!]
\centering
\includegraphics[width=\linewidth]{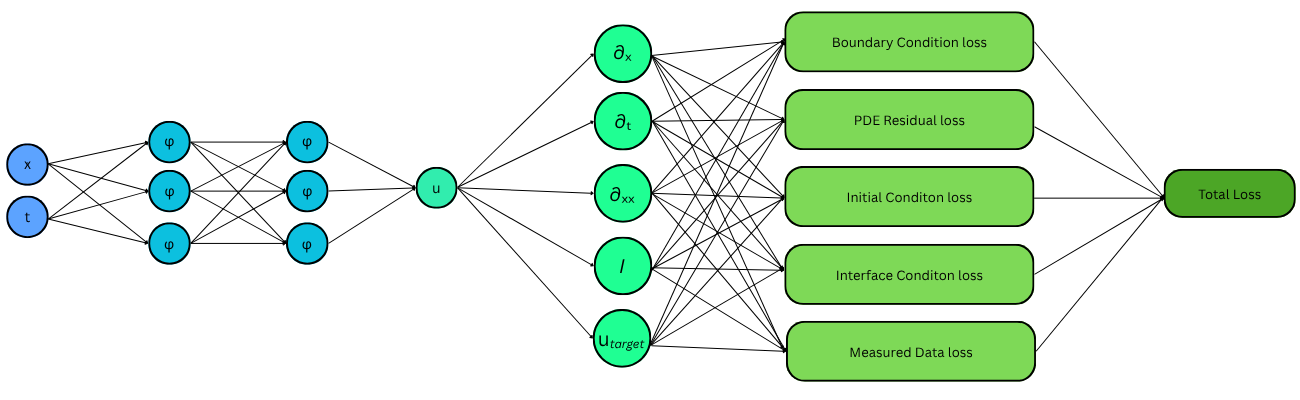}
\caption{Schematic of the XPINN architecture and loss components.}
\label{fig:Figure_1}
\end{figure}

The weights $\lambda_{u_q}$, $\lambda_{F_q}$, $\lambda_{I_q}$, and $\lambda_{IF_q}$ control the contribution of each term. In our experiments, all these weights are set to 1, and for simplicity, their symbols will not be mentioned hereafter. Figure~\ref{fig:Figure_1} shows the overall structure of the XPINN framework, showing the input variables, Neural Networks, and individual loss terms that collectively form the total loss.

\section{Mathematical Formulation}
\label{sec:Mathematical_Formulation}

The mathematical formulation of the proposed framework builds on nonlinear conservation laws to describe two-phase flow in porous media. This section introduces the governing equations, boundary and initial conditions, inter-domain continuity condition, and the associated loss terms that ensure the solution remains consistent with the underlying physics and across computational subdomains.

\subsection{Buckley-Leverett Model}
The mathematical foundation of the Buckley-Leverett model \cite{Buc42} lies in nonlinear scalar conservation laws, which in one spatial dimension take the form
\begin{equation}
\frac{\partial u}{\partial t} + \frac{\partial f(u)}{\partial x} = 0, \quad (x,t) \in \Omega \times (0,T], \label{eq:conservation_law}
\end{equation}

where \(u(x,t)\) represents the conserved scalar quantity and \(f(u)\) is a nonlinear flux function. Such equations are hyperbolic partial differential equations whose solutions can develop steep gradients or discontinuities in finite time, even for smooth initial data. When the flux function is non-convex, i.e., when its second derivative changes sign, the solution structure may involve both shock and rarefaction waves, forming a compound wave \cite{Pet73, LeV92}.

A classical and significant example of this class of problems is the Buckley-Leverett equation, which models the displacement of a non-wetting phase, such as oil, by a wetting phase, such as water, in porous media \cite{Buc42, Azi79}. Typically, water is injected into an oil-saturated formation through injection wells, leading to the displacement of oil towards production wells. Under the assumptions of negligible gravity and capillarity, incompressibility, constant porosity, and one-dimensional flow, the governing conservation equation for the water saturation \(S_w(x,t)\) is written as
\begin{equation}
\frac{\partial S_w}{\partial t} + \frac{\partial f_w(S_w)}{\partial x} = 0, \label{eq:BL_equation}
\end{equation}

where \(f_w(S_w)\) is the fractional flow function of water. Equation~\eqref{eq:BL_equation} is supplemented with the initial and boundary conditions
\begin{equation}
S_w(x,0) = 0, \quad x \in (0,1], \qquad S_w(0,t) = 1, \quad t > 0, \label{eq:BL_IC_BC}
\end{equation}

which represent an initially oil-filled porous medium with unit water saturation imposed at the inlet. The residual and data losses for the Buckley-Leverett equation are defined as follows,

\begin{equation}
L_{F_q}(\theta_q) = \frac{1}{N_{F_q}} \sum_{i=1}^{N_{F_q}} \bigg| \frac{\partial S_{w_{\theta_q}}}{\partial t} + \frac{\partial f_w(S_{w_{\theta_q}})}{\partial x} \bigg|^2, \label{eq:BL_residual_loss}
\end{equation}

\begin{equation}
\begin{aligned}
L_{u_q}(\theta_q) = \frac{1}{N_{u_q}} \sum_{i=1}^{N_{u_q}} \Big( &| S_{w_{\theta_q}}(x_\mathrm{IC}^i, 0) |^2 \\ &+ | S_{w {\theta_q}}(0, t_\mathrm{BC}^i) - 1 |^2 \Big).  \label{eq:BL_data_loss}
\end{aligned}
\end{equation}

\subsection{Flow Nonlinearity}
The nonlinear character of the equation arises from the fractional flow function \( f_w(S_w) \), which determines the fraction of the total fluid flux carried by the water phase. It is derived from the phase mobilities:
\begin{equation}
\lambda_\alpha(S_\alpha) = \frac{k_{r\alpha}(S_\alpha)}{\mu_\alpha}, \qquad \alpha \in \{w, o\},\label{eq:phase_mobility}
\end{equation}
where \( k_{r\alpha} \) are the relative permeabilities and \( \mu_\alpha \) are the viscosities of water (\( w \)) and oil (\( o \)). 

The water fractional flow is then given by
\begin{equation}
f_w(S_w) = \frac{\lambda_w(S_w)}{\lambda_w(S_w) + \lambda_o(S_o)}, \qquad S_o = 1 - S_w. \label{eq:fw_general}
\end{equation}

Different models for the relative permeabilities lead to different shapes of the flux function and, consequently, different wave structures in the solution. However, in most cases, two-phase flow in porous media exhibits nonlinear relative permeabilities, often described by Brooks-Corey type models~\cite{Bro64}. The resulting flux is a nonlinear and non-convex hyperbolic PDE, producing compound solutions with both shocks and rarefactions. Thus, the fractional flow function is given by
\begin{equation}
f_w(S_w) = \frac{S_w^2}{S_w^2 + M (S_o)^2}. \label{eq:fw_specific}
\end{equation}

\subsection{Jump Condition}
To ensure physical consistency across neighboring Sub-Nets, a jump condition is introduced based on the Rankine--Hugoniot condition~\cite{Ran70, Hug87}. This condition enforces that the shock speed $s$ equals the ratio of the flux jump to the state jump across the interface points $(x_{I_q}^i, t_{I_q}^i)$:
\begin{equation}
s = \frac{f_w(S_{w_{\theta_{q^+}}}^i) - f_w(S_{w_{\theta_q}}^i)}{S_{w_{\theta_{q^+}}}^i - S_{w_{\theta_q}}^i},
\label{eq:jump_condition}
\end{equation}

In implementation, the fluxes and solution values of adjacent Sub-Nets are evaluated at the shared interface points, and a small constant $\epsilon = 10^{-8}$ is added for numerical stability. The resulting loss term is defined as
\begin{equation}
L_R(\theta_q) = \frac{1}{N_{I_q}} \sum_{i=1}^{N_{I_q}} 
\left|
\frac{f_w(S_{w_{\theta_{q^+}}}^i) - f_w(S_{w_{\theta_q}}^i)}{S_{w_{\theta_{q^+}}}^i - S_{w_{\theta_q}}^i + \epsilon} - s
\right|^2,
\label{eq:rankine_loss}
\end{equation}

which penalizes deviations from the Rankine--Hugoniot condition, thereby enforcing residual (flux) across subdomain interfaces.


\section{Computational Experiments}
\label{sec:Computational_Experiments}
This section presents the computational setup and numerical experiments performed using the proposed XPINN framework, followed by a comparison against standard PINN and its variants, including PINN with artificial viscosity, PINN with Welge construction, and PINN with the Oleinik entropy condition.

\subsection{Results and Analysis of XPINNs}
\subsubsection{XPINN With Interface Residual}
\label{sec:xpinn_with_interface_residual}
\begin{figure}[H]
\centering
\includegraphics[width=\linewidth]{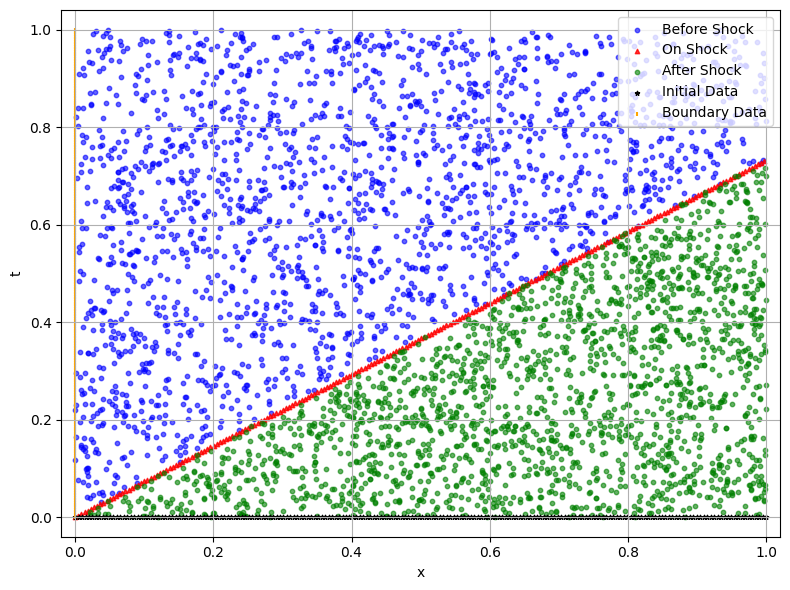}
\caption{Training Data Distribution for XPINN.}
\label{fig:Figure_2}
\end{figure}

\begin{figure}[H]
\centering
\includegraphics[width=\linewidth]{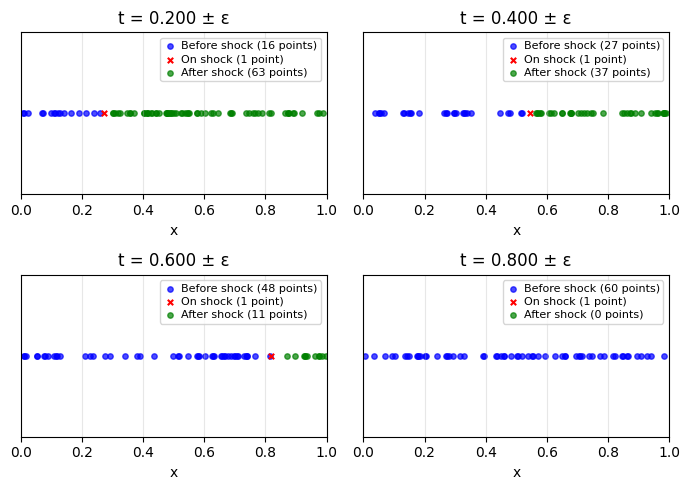}
\caption{Dynamic Domain Decomposition in Space and Time, here $\epsilon = 0.007$.}
\label{fig:Figure_3}
\end{figure}

Here, we show the results of XPINN obtained using Equation \ref{eq:total_loss}. The domain is dynamically decomposed in both space and time into two subdomains—pre-shock and post-shock regions—to accurately capture the evolving discontinuity and distinct flow characteristics across the shock front. The model is implemented for the one-dimensional Buckley-Leverett problem over $x \in [0,1]$ and $t \in [0,1]$ with mobility ratio $M = 2$. The dataset includes 200 points for the initial condition at $t=0$, 200 for the boundary at $x=0$, 2000 collocation points each for the pre- and post-shock regions, and 200 points along the shock interface connecting them, as shown in Figure~\ref{fig:Figure_2}. Moreover, the computational domain is decomposed dynamically so that the subdomains evolve with the moving shock. For each spatial and temporal instance, a single collocation point along the shock interface is maintained, while multiple points are assigned to the pre-shock and post-shock regions. As the shock propagates, the collocation points are updated accordingly, ensuring that the data distribution adapts to the changing flow structure, as illustrated in Figure~\ref{fig:Figure_3}. Since the interface in this case corresponds to the shock region (a discontinuity), the Interface Average term of Eq.~\ref{eq:total_loss} is not enforced, consistent with the reasoning provided in Section~\ref{sec:loss_functions_and_interface_conditions_for_xpinns}.

\begin{figure}[H]
\centering
\includegraphics[width=\linewidth]{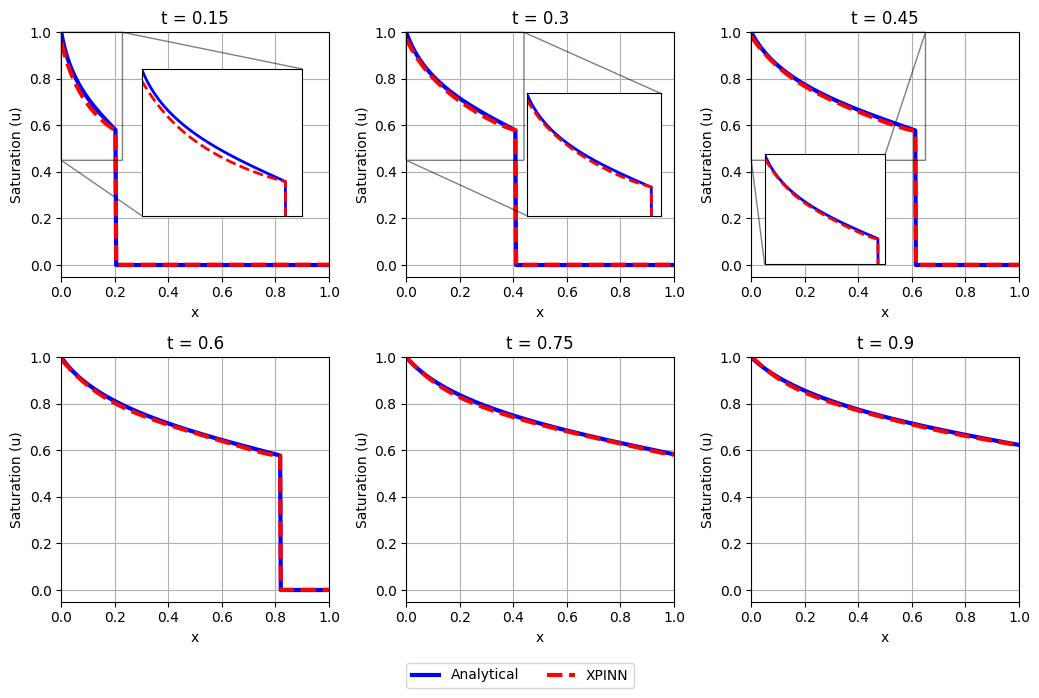}
\caption{Model results for Architecture~1.}
\label{fig:Figure_4}
\end{figure}

\begin{figure}[H]
\centering
\includegraphics[width=\linewidth]{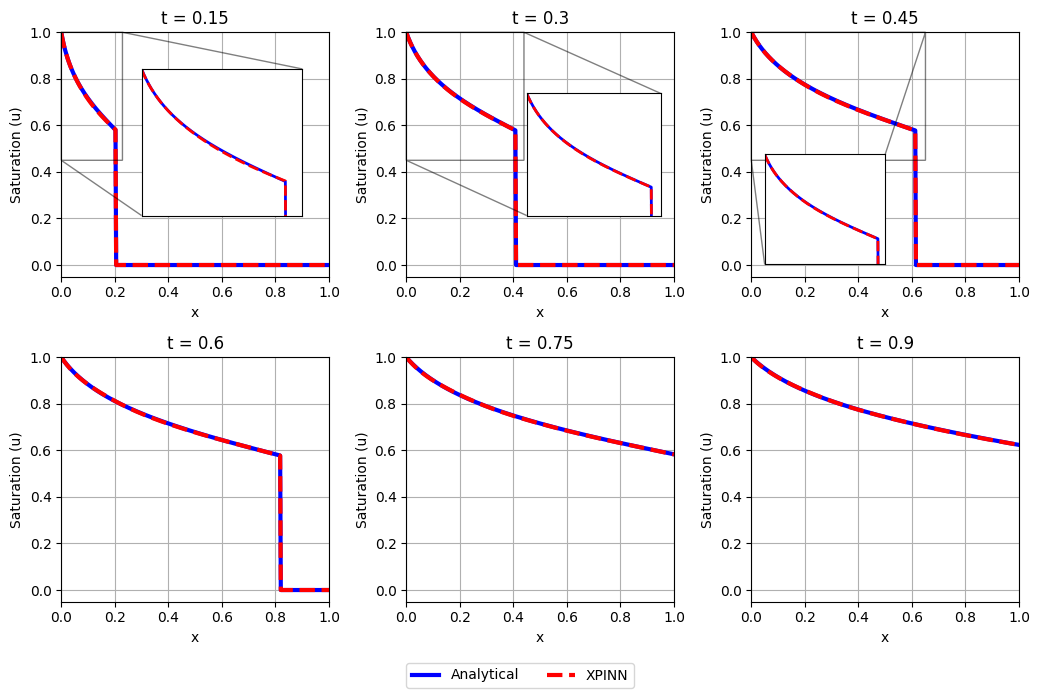}
\caption{Model results for Architecture~2.}
\label{fig:Figure_5}
\end{figure}

We begin with a simple architecture in which the neural networks for subdomain 1 and subdomain 2 each contain only a single hidden layer with ten neurons. Both models use the same learning rate and employ the Tanh function as a locally adaptive activation function, which enhances the network's learning capacity, particularly during the early stages of training. The detailed configuration is summarized in Table~\ref{tab:Table_1}, and the corresponding result is presented in Figure~\ref{fig:Figure_4}. It shows a considerable accuracy overall, with only minor deviations observed during the early training phase (see the zoomed region in Figure~\ref{fig:Figure_4}). To further refine the performance, a slightly complex architectures were explored, as summarized in Table~\ref{tab:Table_2} and results are presented in Figure~\ref{fig:Figure_5}.

\begin{table}[h!]
\centering
\caption{Network configuration (Architecture~1).}
\label{tab:Table_1}
\renewcommand{\arraystretch}{1.4}
\begin{tabular}{lcc}
\toprule
Subdomain & Pre-shock region & Post-shock region \\
\midrule
\# Hidden Layers & 1 & 1 \\
\# Neurons & 10 & 10 \\
Learning Rate & $10^{-3}$ & $10^{-3}$ \\
Optimizer & Adam & Adam \\
Adaptive Activation Function & Tanh & Tanh \\
Minimum Loss & $7.51\times10^{-3}$ & $6.34\times10^{-5}$ \\
Total Minimum Loss & \multicolumn{2}{c}{$7.57\times10^{-3}$} \\
\# Epochs & \multicolumn{2}{c}{5,000} \\
\bottomrule
\end{tabular}
\end{table}

\begin{table}[h!]
\centering
\caption{Network configuration (Architecture~2).}
\renewcommand{\arraystretch}{1.4}
\label{tab:Table_2}
\begin{tabular}{lcc}
\toprule
Subdomain & Pre-shock region & Post-shock region \\
\midrule
\# Hidden Layers & 2 & 1 \\
\# Neurons & 30, 20 & 10 \\
Learning Rate & $10^{-3}$ & $10^{-3}$ \\
Optimizer & Adam & Adam \\
Adaptive Activation Function & Tanh & Tanh \\
Minimum Loss & $1.53\times10^{-3}$ & $1.58\times10^{-5}$ \\
Total Minimum Loss & \multicolumn{2}{c}{$1.55\times10^{-3}$} \\
\# Epochs & \multicolumn{2}{c}{5,000} \\
\bottomrule
\end{tabular}
\end{table}

\begin{figure}[H]
\centering
\includegraphics[width=\linewidth]{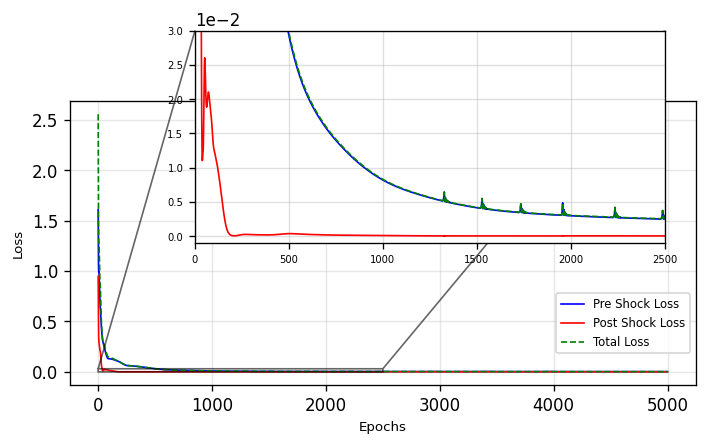}
\caption{Convergence behavior of the losses for Architecture~2.}
\label{fig:Figure_6}
\end{figure}

We also show the convergence behavior of the losses during training, Figure~\ref{fig:Figure_6}. The Pre Shock Loss and Post Shock Loss both decrease rapidly in the initial epochs and stabilize as training progresses. The pre-shock loss remains higher than the post-shock loss since, after the shock, the solution primarily consists of zeros, resulting in smaller residuals. Consequently, the total loss (sum of both losses) is dominated by the pre-shock loss, reflecting that most of the error originates from the region before the shock.

\subsubsection{XPINN Without Interface Residual}
\begin{figure}[H]
\centering
\includegraphics[width=\linewidth]{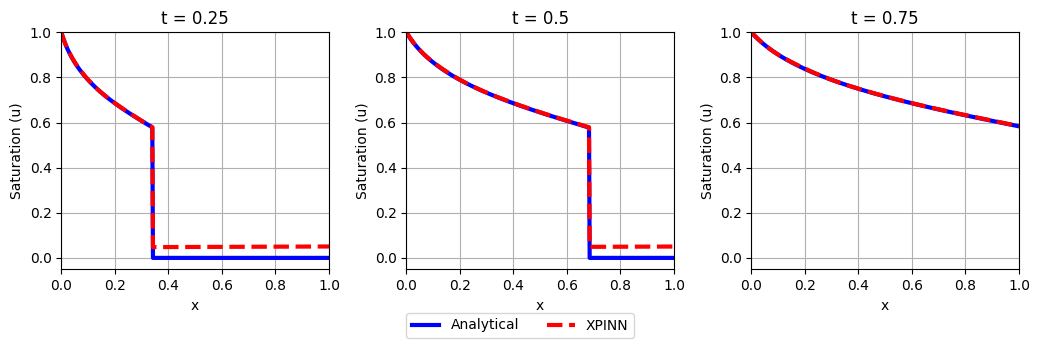}
\caption{Model results for Architecture~2 without Residual Interface term.}
\label{fig:Figure_7}
\end{figure}

Here, we present the results obtained using Equation~\ref{eq:total_loss} without enforcing the interface residual. XPINN without the interface residual is similar to a standard PINN, with the main difference being that two separate PINNs are used—one for each subdomain—instead of a single PINN for the entire domain. All other settings are the same as those described in Section~\ref{sec:xpinn_with_interface_residual} (with interface). Since the interface corresponds to a discontinuity, the average interface term is not applied as explained earlier. Figure~\ref{fig:Figure_7} shows the results for architecture 2 with the hyperparameters summarized earlier in Table~\ref{tab:Table_2}.

As seen in Figure~\ref{fig:Figure_7}, XPINN without residual interface accurately captures the shock location without the residual interface term, though post-shock saturation does not reach zero. We experimented with different values of $M$ (from 0.5 to 30) and observed similar results. So, we deem that the use of separate subnetworks on each subdomain introduces enough representational freedom for a steep interface to form, allowing the model to approximate the correct shock location even without the Rankine-Hugoniot constraint. However, this accuracy is limited: without the residual interface, the interface states are not physically anchored, and the post-shock value in subnetwork 2 remains slightly above zero. This occurs because the solution starts from the interface value, and the PDE residual in the flat post-shock region cannot push it toward the exact physical value: any constant in this region satisfies the PDE locally, so the residual is near zero and the optimizer receives almost no gradient to correct the output. Table~\ref{tab:Table_3} provides a detailed mathematical illustration. Therefore, the residual interface term is essential not for locating the shock but for enforcing the correct left and right states across it, which sets the post-shock value to the physically accurate zero-saturation flat region.

\begin{table}[H]
\centering
\caption{Post-shock region ($u = C$, a constant = 0) comparison without and with interface residual}
\renewcommand{\arraystretch}{2.2}
\label{tab:Table_3}
\begin{tabular}{>{\centering\arraybackslash}p{0.25\textwidth}>{\centering\arraybackslash}p{0.325\textwidth}>{\centering\arraybackslash}p{0.325\textwidth}}
\toprule
\textbf{Component} & \textbf{Without Interface Residual} & \textbf{With Interface Residual} \\
\midrule
Field Solution & $u(x,t) = C$ & $u(x,t) = C$ \\
Spatial Derivative & $\dfrac{\partial u}{\partial x} = 0$ & $\dfrac{\partial u}{\partial x} = 0$ \\
Time Derivative & $\dfrac{\partial u}{\partial t} = 0$ & $\dfrac{\partial u}{\partial t} = 0$ \\
Flux Derivative & $\dfrac{\partial f(u)}{\partial x} = f'(C)\, \dfrac{\partial u}{\partial x} = 0$ & $\dfrac{\partial f(u)}{\partial x} = f'(C)\, \dfrac{\partial u}{\partial x} = 0$ \\
PDE Residual & $R_{\text{PDE}} = \dfrac{\partial u}{\partial t} + \dfrac{\partial f(u)}{\partial x} = 0$ & $R_{\text{PDE}} = \dfrac{\partial u}{\partial t} + \dfrac{\partial f(u)}{\partial x} = 0$ \\
Interface Residual & $R_{\text{interface}} = 0 $ & $R_{\text{interface}} = (u_{\text{post-shock}} - u_{\text{interface}}) \neq 0$ \\
Total Residual & $R_{\text{total}} = 0 $ & $R_{\text{total}} = R_{\text{PDE}} + R_{\text{interface}} \neq 0$ \\
Interpretation & \textit{The PDE is satisfied exactly even if $C \neq 0$. Optimizer gets almost no gradient to correct post-shock value.} & \textit{The interface residual creates a gradient at the shock interface, which propagates into the post-shock region and forces the network toward the correct value ($C=0$).} \\
\bottomrule
\end{tabular}
\end{table}

\subsection{Comparision of XPINN with Standard PINN and its variants}
\begin{figure}[H]
\centering
\includegraphics[width=\linewidth]{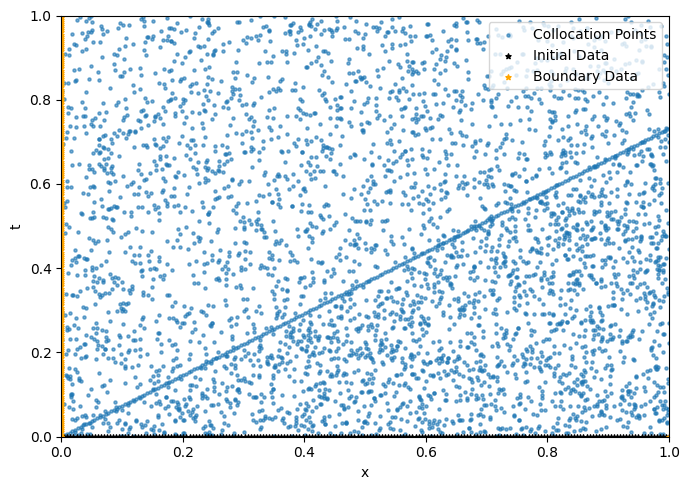}
\caption{Training Data Distribution for Standard PINN.}
\label{fig:Figure_8}
\end{figure}

In this section, we compare the results of the XPINN approach with the Standard PINN and its variants, including PINN with artificial viscosity (diffusivity term), PINN with Welge construction, and PINN with the Oleinik entropy condition. To ensure a fair comparison, the same dataset and training points used for XPINN (Section~\ref{sec:xpinn_with_interface_residual}) are employed for all Standard PINN variants. Before presenting the results and associated mathematical formulations, we first illustrate the data distribution used for the Standard PINN. The setup is identical to that described in the XPINN section, except that here we consider a single unified dataset rather than separate subdomains, as shown in Figure~\ref{fig:Figure_8}.

\subsubsection{Standard PINN}
Equation~\ref{eq:total_loss} shows the Standard PINN part which is

\begin{equation}
J(\theta_q) =
\underbrace{L_{u_q}(\theta_q)}_{\text{Data}}
+ \underbrace{L_{F_q}(\theta_q)}_{\text{Residual}}
\end{equation}
where, $L_{u_q}$ and $L_{F_q}$ are the same as defined by Equation~\ref{eq:data_loss} and~\ref{eq:PDE_res}.

\subsubsection{PINN with Diffusivity Term}
A second-order artificial viscosity (diffusivity) term is incorporated into the residual \cite{Fuk20}. Accordingly, the residual term in Equation~\ref{eq:BL_residual_loss} is modified as:
\begin{equation}
L_{F_q}(\theta_q) = \frac{1}{N_{F_q}} \sum_{i=1}^{N_{F_q}} 
\bigg| \frac{\partial S_{w_{\theta_q}}}{\partial t} 
+ \frac{\partial f_w(S_{w_{\theta_q}})}{\partial x} 
- \epsilon \frac{\partial^2 S_{w_{\theta_q}}}{\partial x^2} \bigg|^2,
\label{eq:BL_residual_loss2}
\end{equation}
where $\epsilon > 0$ denotes the diffusion coefficient. Based on \cite{Fuk20}, $\epsilon$ is set to $2.5 \times 10^{-3}$.

\subsubsection{PINN with Welge Construction}
Using the Welge construction to enforce the convex hull of the fractional flow \cite{Fra201}, the modified fractional flow function is expressed as:
\begin{equation}
\tilde{f}(S_w) =
\begin{cases} 
0, & S_w = 0 \\[1mm]
\dfrac{S_w}{f(S^*)}, & 0 < S_w \leq S^* \\[1mm]
f(S_w), & S_w > S^*
\end{cases}
\end{equation}

\subsubsection{PINN with Oleinik Entropy Condition}
The Oleinik entropy condition is applied to maintain physically admissible shock solutions \cite{Dia21}, leading to the following convex-hull representation of the fractional flow:
\begin{equation}
\tilde{f}(S_w) =
\begin{cases} 
\sigma S_w, & 0 \le S_w < S^* \\[1mm]
f_w(S_w), & S_w \ge S^*
\end{cases}
\end{equation}

where, $S^*$ denotes the saturation at the shock front, and $\sigma$ represents the corresponding shock speed, which is given by

\begin{equation}
\sigma = \frac{f(S^*)}{S^*}.
\end{equation}

\begin{figure}[H]
\centering
\includegraphics[width=\linewidth]{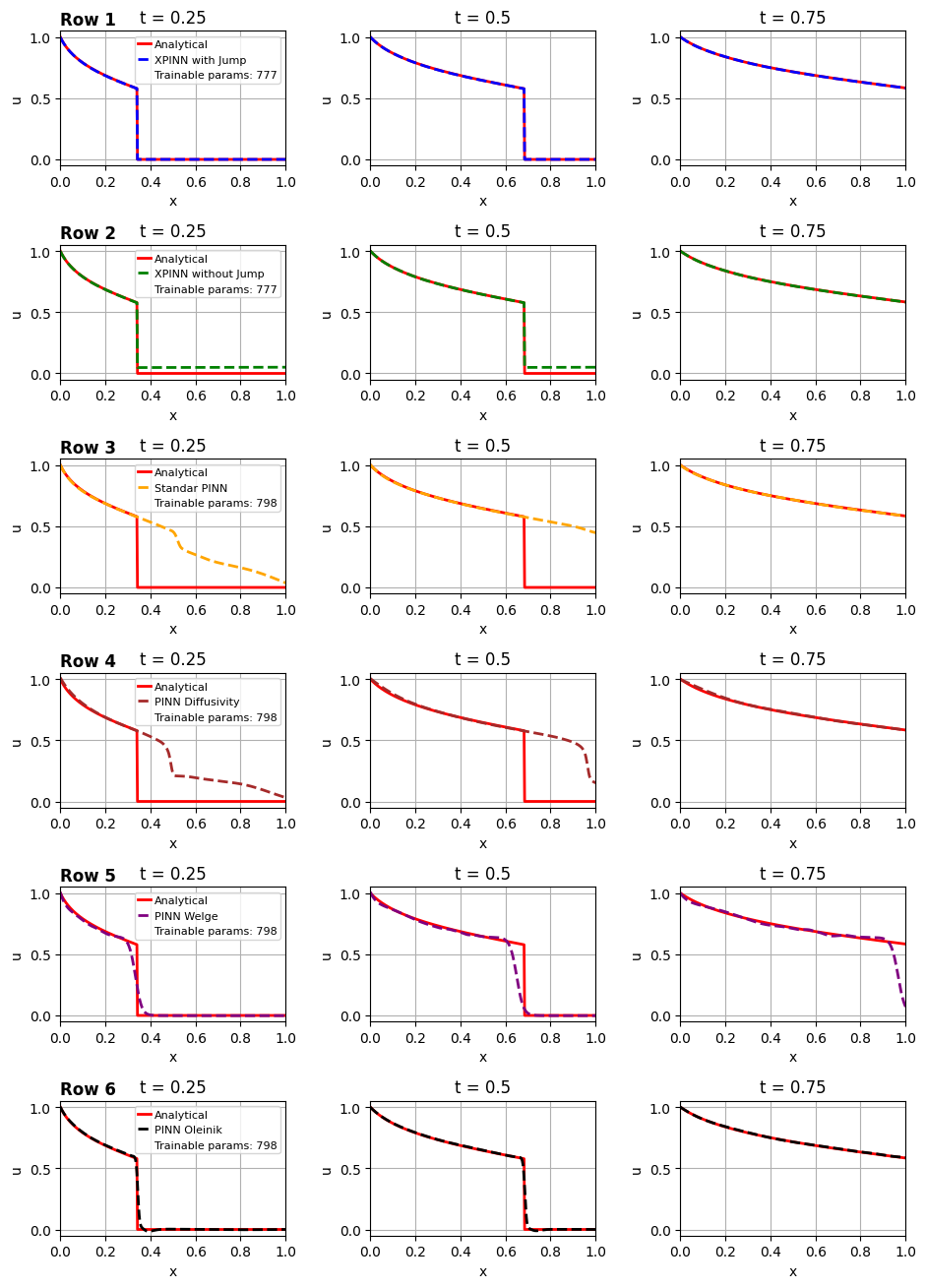}
\caption{Comparision of XPINN with Standard PINN and its variants for $M = 2$.}
\label{fig:Figure_9}
\end{figure}

Building upon these formulations, we conducted comprehensive experiments comparing all methods for the case of $M = 2$. It is important to note that all models were configured with approximately the same number of trainable parameters and identical hyperparameters, as detailed in Table~\ref{tab:Table_4}, ensuring a fair comparison across methods. The predictive results presented in Figure~\ref{fig:Figure_9} and as reported in prior studies, the Standard PINN approach (Row~3) fails to accurately capture the solution and post shock saturation. Similarly, the PINN with artificial diffusivity (Row~4) demonstrates limited improvement and fails to resolve the sharp interface adequately. While \cite{Fuk20} reported successful predictions using the L-BFGS optimizer, our comparative analysis employs the Adam optimizer consistently across all methods. The PINN variants incorporating Welge construction (Row~5) and Oleinik entropy condition (Row~6) yield physically plausible solutions with moderate accuracy, though their performance remains inferior to the XPINN method (Row~1), which demonstrates superior resolution of the shock front.

\begin{table}[h!]
\centering
\caption{Model configuration and hyperparameters for XPINN and PINN variants.}
\renewcommand{\arraystretch}{1.6}
\label{tab:Table_4}
\begin{tabular}{
>{\centering\arraybackslash}m{0.18\textwidth}
>{\centering\arraybackslash}m{0.14\textwidth}
>{\centering\arraybackslash}m{0.14\textwidth}
>{\centering\arraybackslash}m{0.14\textwidth}
>{\centering\arraybackslash}m{0.14\textwidth}
>{\centering\arraybackslash}m{0.14\textwidth}
}
\toprule
Component & \shortstack[c]{XPINN \\ {\scriptsize (2 sub-domains)}} & Standard PINN & PINN Diffusivity & PINN Welge & PINN Oleinik \\
\midrule
\# Hidden Layers & 2 and 1 & 2 & 2 & 2 & 2 \\
\# Neurons & (30, 20) and (10) & 30, 22 & 30, 22 & 30, 22 & 30, 22 \\
\# Epochs & 5,000 & 5,000 & 5,000 & 5,000 & 5,000 \\
Total Trainable Parameters & 777 & 798 & 798 & 798 & 798 \\
Total Collocation Points & 4,600 & 4,600 & 4,600 & 4,600 & 4,600 \\
Learning Rate & $10^{-3}$ & $10^{-3}$ & $10^{-3}$ & $10^{-3}$ & $10^{-3}$ \\
Optimizer & Adam & Adam & Adam & Adam & Adam \\
Adaptive Activation Function & Tanh & Tanh & Tanh & Tanh & Tanh \\
\bottomrule
\end{tabular}
\end{table}

\clearpage

\begin{figure}[H]
\centering
\includegraphics[width=\linewidth]{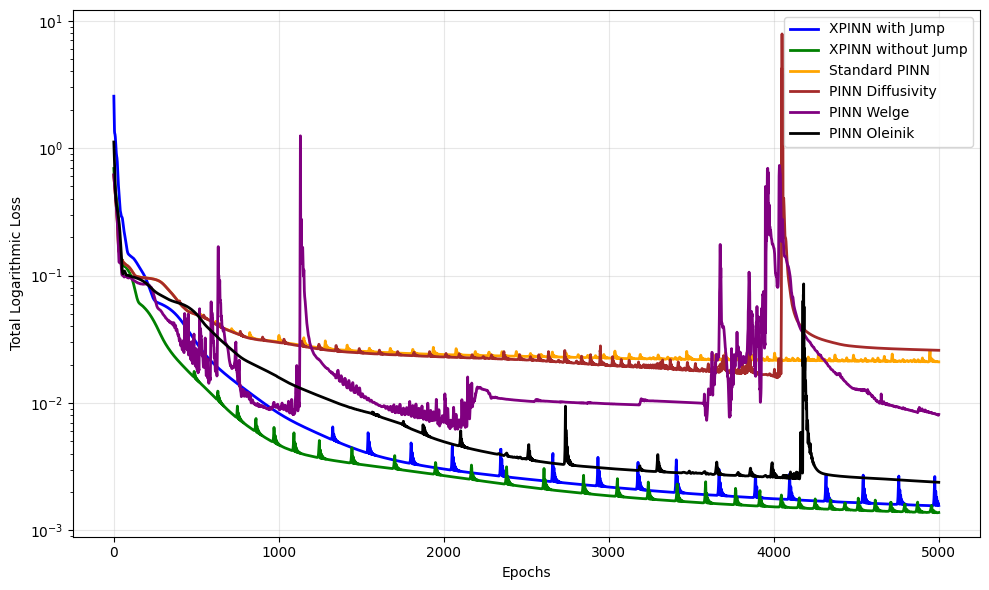}
\caption{Convergence behaviour of all methods for $M = 2$.}
\label{fig:Figure_10}
\end{figure}

\begin{figure}[H]
\centering
\includegraphics[width=\linewidth]{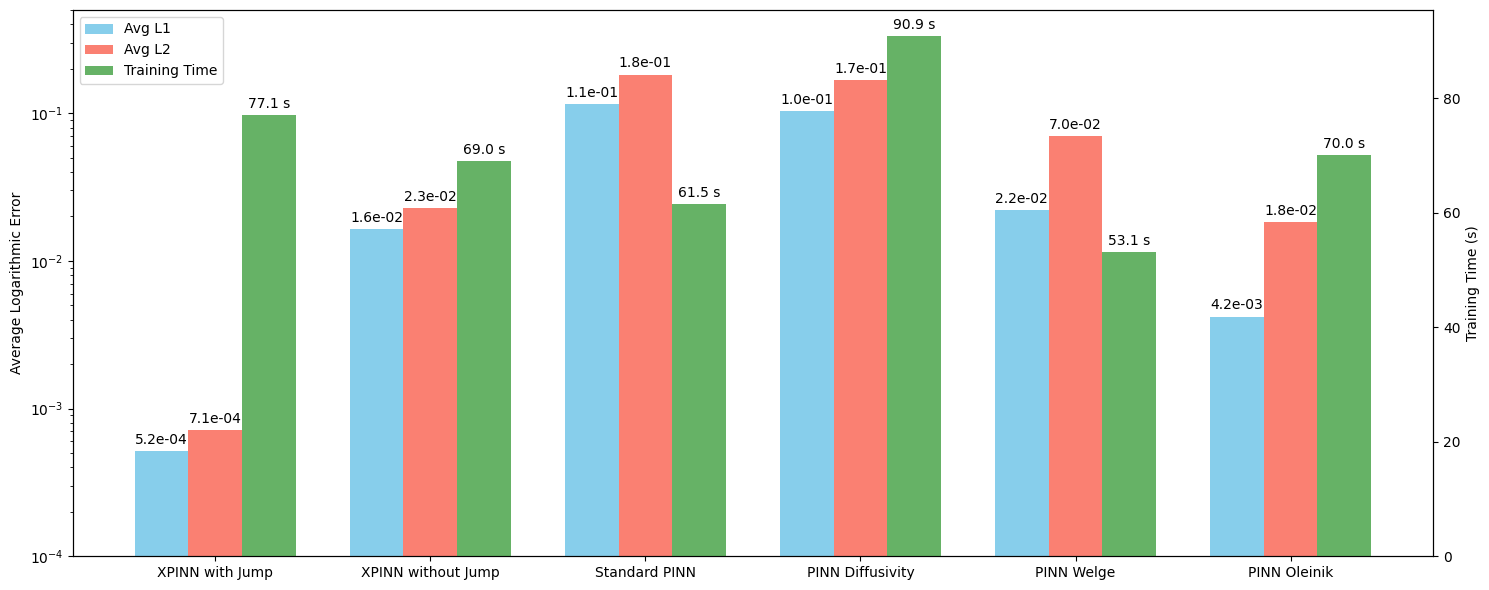}
\caption{Quantitative metrics of all methods for $M = 2$.}
\label{fig:Figure_11}
\end{figure}

The convergence behavior for $M = 2$, illustrated in Figure~\ref{fig:Figure_10}, reveals that XPINN exhibits significantly smoother and more stable optimization dynamics compared to other methods, which display substantial oscillations during training. This enhanced convergence property is further substantiated by the quantitative metrics presented in Figure~\ref{fig:Figure_11}, which compares all methods using L1 and L2 error norms on test dataset alongside training time. Although XPINN requires marginally more computational time, it achieves the lowest error metrics among all compared methods, demonstrating an excellent trade-off between accuracy and computational cost.

\clearpage

Notably, for the more challenging case of $M = 30$ (Figure~12), the performance gap becomes even more pronounced. While the Oleinik entropy condition and other PINN variants fail to produce accurate predictions under these more demanding conditions, the XPINN approach maintains robust performance, successfully resolving the sharper shock front and demonstrating the method's scalability and robustness across different mobility ratio scenarios. The hyperparameters used in this study are the same as those shown in Table~\ref{tab:Table_4}.

\begin{figure}[H]
\centering
\includegraphics[width=\linewidth]{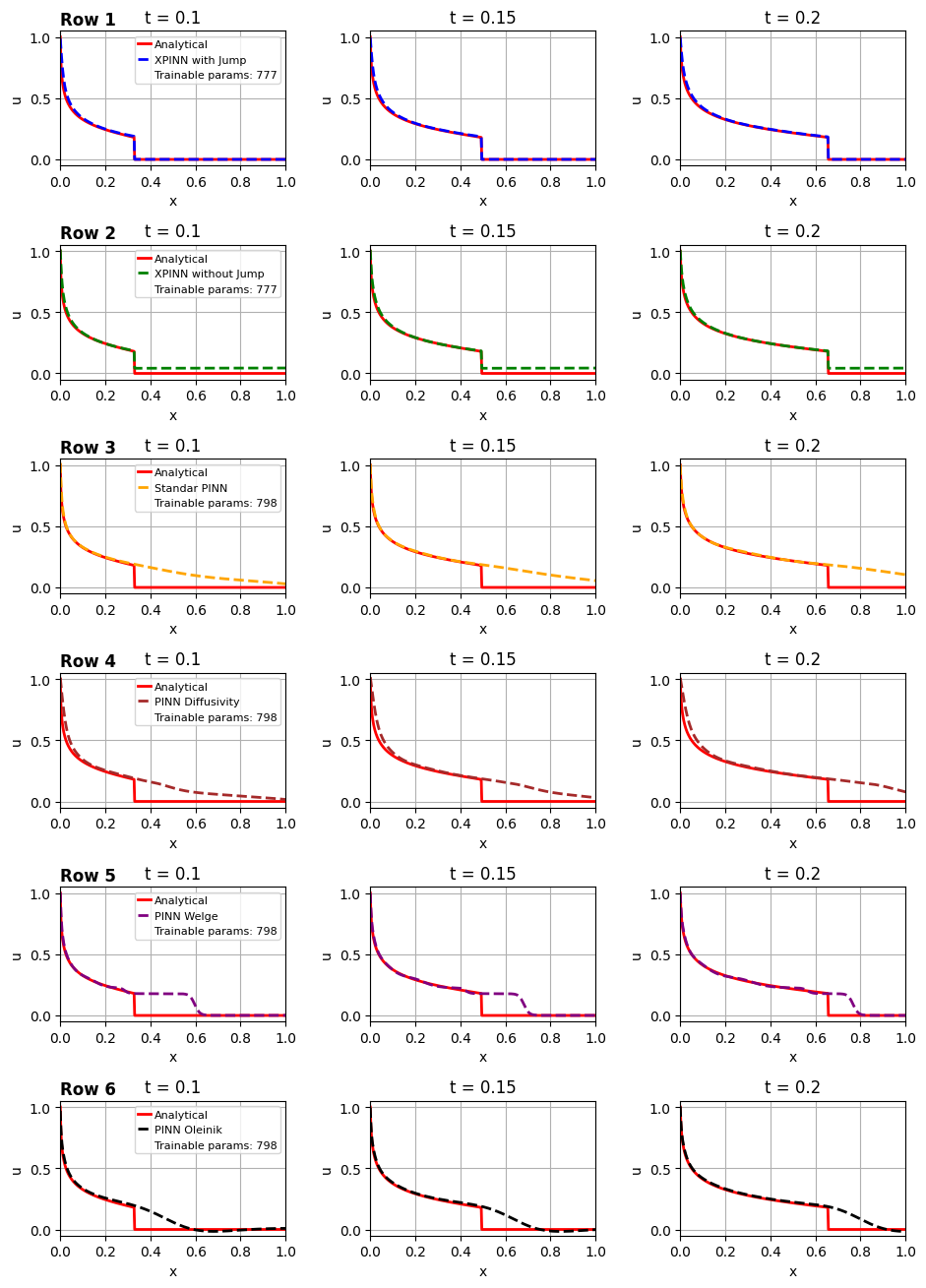}
\caption{Comparision of XPINN with Standard PINN and its variants for $M = 30$.}
\label{fig:Figure_12}
\end{figure}


\section{Conclusion and Future Work}
\label{sec:Conclusion_and_Future_Work}

This work applied the Extended Physics-Informed Neural Network (XPINN) framework to the Buckley-Leverett equation with a nonconvex flux, demonstrating that dynamic domain decomposition in space-time—partitioning the domain into evolving pre-shock and post-shock regions—enables each subnetwork to learn localized solution structure and collectively represent sharp saturation fronts. Enforcing the Rankine-Hugoniot jump condition along the moving interface provided a physically consistent coupling between subnetworks and allowed the model to capture accurate behavior without introducing artificial diffusion or entropy constraints. 

We compared the results of the XPINN approach with the Standard PINN and its variants, including PINN with artificial viscosity (diffusivity term), PINN with Welge construction, and PINN with the Oleinik entropy condition. Across all cases and mobility ratios, XPINN consistently outperformed the other methods, accurately resolving sharp fronts and capturing the correct physical behavior, while the other PINN variants struggled under increasingly challenging conditions. Significantly, XPINN achieved this using the simpler Adam optimizer, whereas some PINN variants required more complex or higher-order optimization strategies, such as L-BFGS, to reach comparable accuracy.

Looking forward, we plan to refine the training strategy by employing separate training loops and stopping criteria per subdomain so that regions with simpler dynamics can converge with fewer epochs while more challenging regions receive additional training. Furthermore, integrating the Residual-based Adaptive Refinement (RAR) strategy within each subdomain can dynamically allocate collocation points in regions with high residual error, further enhancing local accuracy and training efficiency. We will also extend the methodology to two- and three-dimensional configurations and adapt the dynamic domain decomposition to those settings, with particular attention to heterogeneous reservoirs and realistic boundary conditions. Finally, we aim to translate the approach into practical water-flooding studies—validating XPINN on field-scale, two-phase displacement scenarios to assess its capability for history matching, forward prediction, and decision support in reservoir engineering.


\section{Declaration of generative AI in the manuscript preparation process}
During the preparation of this work the authors used ChatGPT in order to improve the writing of the manuscript. After using this tool, the authors reviewed and edited the content as needed and take full responsibility for the content of the published article.


\input{paper.bbl}

\end{document}